\documentclass{ws-ijufks}

\begin{document}

\markboth{X.C. Guan and Y.M. Li}{Continuity in Information Algebras}

%

%

\title{CONTINUITY IN INFORMATION ALGEBRAS: A SURVEY ON THE RELATIONSHIP BETWEEN TWO TYPES OF INFORMATION ALGEBRAS}

\author{XUECHONG GUAN\footnote{Work Address: College of Mathematic Science, Xuzhou Normal University,
Xuzhou, 221116, China.}}

\address{College of Mathematics and Information Science\\
Shaanxi Normal University\\
Xi'an, 710062, China\\
guanxc@foxmail.com}

\author{YONGMING LI}

\address{College of Computer Science\\
Shaanxi Normal University\\
Xi'an, 710062, China\\
liyongm@snnu.edu.cn}

\maketitle

\begin{history}
\received{24 June 2010}
\revised{(revised date)}
\end{history}

\begin{abstract}
In this paper, the continuity and strong continuity in domain-free information algebras and labeled information algebras are
introduced respectively. A more general concept of continuous function which is defined between two
domain-free continuous information algebras is presented.
It is shown that, with the operations combination and focusing, the set of all continuous functions between two domain-free
s-continuous information algebras forms a new s-continuous information algebra.
By studying the relationship between domain-free information algebras and labeled information algebras, it is demonstrated that
they do correspond to each other on s-compactness.
\end{abstract}

\keywords{domain-free continuous information algebra; labeled continuous information algebra; continuous function;
compactness.}

\section{Introduction}\label{intro}
Inference under uncertainty is a common problem in the real world.
Thus, for pieces of information from different sources, there always exist two fundamental aspects that to combine
information and to exact information on a designated domain.
Based on the above consideration, the valuation-based system (VBS) was first introduced by Shenoy. \cite{VBS}
Kohlas, in Ref. 2, has exactly introduced the concept of information algebra.
We can see that information algebra is an algebraic structure links up with local computation and inference for treating
uncertainty or, more generally, information and knowledge.
It gives a basic mathematical model for describing the modes of information processing.
Recent studies \cite{Information} \cite{Lecture} \cite{Semiring} showed that
the framework of information algebra covers many instances from constraint systems, Bayesian networks, Dempster-Shafer
belief functions to relational algebra, logic and etc.

In view of the feasibility of computer processing information,
Kohlas gave the notions of domain-free compact information algebra and labeled compact information algebra successively
in the study of representation of information algebras. In the light of the previous conclusions,
we know that there exists a correspondence between domain-free information algebra and labeled information algebra,
that is, from a domain-free information algebra, we can construct its associated labeled information algebra, and vice versa.
But, labeled compact information algebras introduced in Ref. 3 do not necessarily lead to domain-free compact information algebras,
as we have seen in the example of cofinite sets.\cite{Lecture} It naturally raises a question that
whether we can present an improved definition of labeled compact information algebra such that
its associated domain-free information algebra is compact and strong compact respectively.
Accordingly, in this paper, we redefine the notions of labeled continuous
information algebra and domain-free continuous information algebra respectively. Obviously
compact information algebras defined in the previous literature can be seen as a special case of continuous information
algebras. As a result, the conclusions which are obtained in continuous information algebras are also more extensive.
It should be noted that the definitions in this paper are different from continuous information algebra presented in Ref. 3.
The main difference exists in the characterization of continuity in labeled information algebras.

The main work of this paper is as follows. Basing on the notions of continuous information algebras,
we discuss the correspondence of continuity in labeled information algebras and domain-free
information algebras. The accordance on s-compactness in information algebras thus follows from the conclusions on continuity,
that is, a labeled s-compact information algebra induces its associated domain-free s-compact information algebra and, in turn,
a domain-free s-compact information algebra induces its associated labeled s-compact information algebra too.
We also present the equivalent statements of some definitions introduced in the paper,
and especially study the properties of function spaces of domain-free continuous information algebras.

The paper is organized as below. Section 2 briefly reviews some basic notions on information algebra.
In Section 3 we introduce the concept of domain-free continuous information algebra and discuss the property of
continuous function spaces. In Section 4 we give the notion of labeled continuous information algebra and focus on the
relationship on continuity and compactness in labeled information algebras and domain-free information algebras.

\section{Preliminaries}

Let's recall some definitions and notations in the theory of information algebra.
For a full introduction, we can refer to Ref. 2-6.
In this study, the fundamental elements of an information algebra are valuations.
In general, a valuation is a function that provides possible elements of a field for variables.
Here a valuation represents some knowledge and information which may be a function, tuple or symbol.

Let $\Phi$ be a set of valuations, and let $D$ be a lattice. Suppose there are three operations defined:

1.Labeling: $\Phi \rightarrow D; \phi\mapsto d(\phi)$, where $d(\phi)$ is called the domain of $\phi$.
Let $\Phi_s$ denote the set of all valuations with domain $s$.

2.Combination: $\Phi\times \Phi \rightarrow \Phi; (\phi,\psi)\mapsto \phi\otimes\psi$,

3.Marginalization: $\Phi\times D \rightarrow \Phi; (\phi,x)\mapsto \phi^{\downarrow x}$, for $x\leq d(\phi)$.

If the system $(\Phi,D)$ satisfies the following axioms, it is called a labeled information algebra:

1.Semigroup: $\Phi$ is associative and commutative under combination.
For all $s\in D$ there is an element $e_s$ with $d(e_s)=s$ such that for all
$\phi\in \Phi$ with $d(\phi)=s, e_s\otimes \phi=\phi$. Here $e_s$ is called a neutral element of $\Phi_s$.

2. Labeling: For $\phi,\psi \in \Phi$, $d(\phi\otimes\psi) = d(\phi)\vee d(\psi)$.

3. Marginalization: For $\phi\in \Phi, x\in D, x\leq d(\phi), d(\phi^{\downarrow x})=x$.

4. Transitivity: For $\phi\in \Phi$ and $x\leq y\leq d(\phi),
(\phi^{\downarrow y})^{\downarrow x}=\phi^{\downarrow x}$

5. Combination: For $\phi,\psi \in \Phi$ with $d(\phi)=x, d(\psi)=y,
(\phi\otimes\psi)^{\downarrow x}=\phi\otimes\psi^{\downarrow x\wedge y}$.

6. Stability: For $x,y\in D, x\leq y$, $e_y^{\downarrow x}=e_x$.

7. Idempotency: For $\phi\in \Phi$ and $x\in D, x\leq d(\phi)$,
$\phi\otimes \phi^{\downarrow x}=\phi$.

The items putting forward in the above definition are the axiomatic presentations of some basic and logical principles
in the process of handling information. In fact, the algebraic structure shown in the definition
covers many instances from expert systems, constraint systems and possibility theory to
relational algebra and logic. We now look at some examples of labeled information algebras.

\begin{example}\label{example:4}(Constraint System)
{\rm A constraint system \cite{constraint} is a tuple $CS=\langle S, D, V\rangle$,
where $\langle S,+,\times,0,1\rangle$ is a semiring, $V$ is a totally ordered set of variables via ordering $\prec$.
$D$ is a finite set which contains at least two elements, called the domain of variables.
A tuple $\langle S,+,\times,0,1\rangle$ is defined to be a semiring, if it satisfies that
both operations + and $\times$ are commutative and associative, and $\times$ distributes over $+$.
The element $0$ is a unit element of $+$ and an absorbing element of $\times$, $1$ is a unit element of $\times$.
A semiring $S$ is called c-semiring,\cite{Semiring} if it is such that for all $a\in S, a + 1 = 1$.

A tuple $c=\langle def, con\rangle$ is called a constraint over CS, where

(\romannumeral 1) $con \subseteq V$, it is called the type of the constraint, denoted by $d(c)=con$;

(\romannumeral 2) $def: D^{|con|}\rightarrow S$, where $|con|$ is the cardinality of $con$.

Let $C$ denote the set of all constraints over CS. Two operations are defined as follows:

1. Combination $\otimes$: For two constraints $c_1=\langle def_1, con_1\rangle,c_2=\langle def_2, con_2\rangle$,
their combination, written $c_1\otimes c_2$, is the constraint
$\langle def, con\rangle$ with $con=con_1\cup con_2$ and $def: D^{|con|}\rightarrow S$ is defined as:
\begin{center}
$def(x)=def_1(x^{\downarrow con_1})\times def_2(x^{\downarrow con_2}), \forall x\in D^{|con|},$
\end{center}
where $x^{\downarrow con_1}$, called tuple projection, is defined as follows:
Suppose $x = \langle x_1, \cdots, x_k \rangle\in D^{|con|}$,
$con_1 = \{v_1^{'},\cdots, v_m^{'}\} \subseteq con = \{v_1,\cdots, v_k\}$,
where $v_i \prec v_j$ and $v_i^{'}\prec v_j^{'}$ if $i< j$,
then $x^{\downarrow con_1}=\langle t_1^{'}, \cdots, t_m^{'} \rangle$, where $t_i^{'}=x_j$ if $v_i^{'}=v_j$.

2. Projection $\Downarrow$: For a constraint $c=\langle def, con\rangle$, if $I\subseteq con$,
the projection of $c$ over $I$, written $c^{\Downarrow I}$, is the constraint
 $\langle def^{'}, I\rangle$ with
\begin{center}
$def^{'}(x)=\sum\limits_{z\in D^{|con|}: z^{\downarrow I}= x} def(z), \forall x\in D^{|I|}.$
\end{center}

With the three operations combination, projection and type, the system $(C, {\cal P}(V))$ induced by a semiring $S$
is a labeled information algebra if, and only if, the semiring $S$ is such that $a\times(a + b)=a$ for all $a, b\in S$.

In fact, if $S$ satisfies with $a\times(a + b)=a$ for all $a, b\in S$,
then $S$ is a c-semiring with the idempotent operation $\times$.\cite{compact}
Hence $(C, {\cal P}(V))$ is an information algebra.\cite{Semiring}
Conversely, let $(C, {\cal P}(V))$ be an information algebra, a variable $v\in V$,
and $D = \{ y_1, y_2,\cdots,y_n \} (n\geq 2)$.
If $a,b\in S$, we take a constraint $c=\langle def, con\rangle$, where $con=\{v\}$ and
$def: D \rightarrow S$ is defined as $def(y_1)=a, def(y_2)=b$ and $def(y)=0$ for all other $y\in D$.
By the idempotency of information algebra, we have
$c\otimes c^{\Downarrow \emptyset}=c$, then $def(y_1)=def(y_1)\times \sum\limits_{y\in D}def(y)$, i.e., $a \times (a + b) = a$.}
\end{example}

\begin{example}\label{example:3}(soft set)
{\rm Let $U$ be an initial universe set and let $E$ be a set of parameters which usually are initial attributes,
characteristics, or properties of objects in $U$. Let ${\cal P}(U)$ denote the power set of $U$ and $A\subseteq E$.
A pair $(F,A)$ is called a soft set \cite{first}over $U$, where $F$ is a mapping given by $F : A\rightarrow {\cal P}(U)$.
A soft set $(F, A)$ over $U$ is said to be a null soft set, if for all $e\in A, F(e)=\emptyset$.

There are three operations defined:

1. Labeling $d$: For a soft set $(F, A)$, we define $d((F, A))=A$.

2. Projection $\downarrow$ \cite{application}:
If $B\subseteq A$, we define $(F, A)^{\downarrow B}$ to be a soft set $(G,B)$ such that for all $b\in B$, $G(b)=F(b)$.

3. Extended intersection $\sqcap$ \cite{soft}:
The extended intersection of two soft sets $(F, A)$ and $(G, B)$ over a common universe $U$ is the soft set $(H, C)$,
where $C = A \cup B$, and $\forall e \in C$,
$$H(e)=\left\{
          \begin {array}{ll}
          F(e),&{\mbox{if}} \ \ e\in A-B;\\
          G(e),&{\mbox{if}} \ \ e\in B-A;\\
          F(e)\cap G(e),&{\mbox{if}} \ \ e\in A\cap B.
         \end{array}
        \right.$$
We write $(F,A)\sqcap (G,B)=(H,C)$.

We are going to show $({\cal F},{\cal P}(E))$ is an information algebra
with the three operations $d, \downarrow$ and $\sqcap$ defined as above, where ${\cal F}$ is the set of all soft sets over $U$.

1. Semigroup: ${\cal F}$ is associative and commutative under $\sqcap$. The null soft set $(\emptyset, \emptyset)$ is a neutral
element such that $(F,A)\sqcap(\emptyset, \emptyset)=(F,A)$ for all soft set $(F,A)$.

2. The axioms of labeling, marginalization, transitivity and idempotency are clear.

3. Combination: For $(F,A), (G,B)\in {\cal F}$, if $A\subseteq S\subseteq A\cup B$, we need to show
$((F,A) \sqcap (G,B))^{ \downarrow S} = (F,A)\sqcap (G,B)^{\downarrow {S \cap B}}$.
In fact, let $(F,A) \sqcap (G,B)=(H, A\cup B)$ and $(F,A)\sqcap (G,B)^{\downarrow {S \cap B}}=(H^{'}, S)$.
For all $e\in S$, we have
$$H(e)=H^{'}(e)=\left\{
          \begin {array}{ll}
          F(e),&{\mbox{if}} \ \ e\in S\cap(A-B);\\
          G(e),&{\mbox{if}} \ \ e\in S\cap(B-A);\\
          F(e)\cap G(e),&{\mbox{if}} \ \ e\in A\cap B.
          \end{array}
          \right.
          $$
Then $((F,A) \sqcap (G,B))^{ \downarrow S} = (F,A)\sqcap (G,B)^{\downarrow {S \cap B}}$.

Hence $({\cal F},{\cal P}(E))$ is a labeled information algebra.}
\end{example}

Now we look at the notion of domain-free information algebra. A system $(\Psi,D)$ with
two operations defined,

1. Combination: $\Psi\times \Psi \rightarrow \Psi; (\phi,\psi)\mapsto \phi\otimes\psi$,

2. Marginalization: $\Psi\times D \rightarrow \Psi; (\psi,x)\mapsto \psi^{\Rightarrow x} $,

We impose the following axioms on $\Psi$ and $D$, and it is called a domain-free information algebra:

1. Semigroup: $\Psi$ is associative and commutative under combination,
and there is an element $e$ such that for all
$\psi\in \Psi$ with $e\otimes \psi=\psi\otimes e=\psi$.

2. Transitivity: For $\psi\in \Phi$ and $x, y\in D,
(\psi^{\Rightarrow y})^{\Rightarrow x}=\psi^{\downarrow x\wedge y}$.

3. Combination: For $\phi,\psi \in \Psi, x\in D$,
$(\phi^{\Rightarrow x}\otimes\psi)^{\Rightarrow x}=\phi^{\Rightarrow x}\otimes\psi^{\Rightarrow x}$.

4. Support: For $\psi\in \Psi$, there is an $x\in D$ such that $\psi^{\Rightarrow x}=\psi$.

5. Idempotency: For $\psi\in \Psi$ and $x\in D,\psi\otimes \psi^{\Rightarrow x}=\psi$.

For simplicity, information algebra is used as a general name for labeled information algebras and
domain-free information algebras. While, we can judge that whether an information algebra is ``labeled" or not from the context.

An abstract example of domain-free information algebra is given below.
\begin{example}\label{example:2}
{\rm Let $\Phi=[0,1]$, $D=\{0,1\}$. We define the following operations:
\begin{center}
Combination: $\forall \phi,\psi\in \Phi, \phi\otimes\psi=\max\{\phi,\psi\};$\\
Focusing: $\forall \phi\in \Phi$, (\romannumeral 1)$\phi^{\Rightarrow 1}=\phi.$\\
(\romannumeral 2) $\phi^{\Rightarrow 0}=\left\{
          \begin {array}{ll}
          \phi,&{\mbox{if}} \ \ \phi\in[0,\frac{1}{2}];\\
          \frac{1}{2},&{\mbox{if}} \ \ \phi\in[\frac{1}{2},1].
         \end{array}
        \right.$
\end{center}
We have $\phi\ll \psi$ if, and only if, $\phi<\psi$ or $\phi=\psi=0$.

It's clear that the element $0$ is a neutral element and the axioms of transitivity and support are correct.
Since $\phi^{\Rightarrow x}\leq \phi$ for all $\phi\in\Phi, x\in D$, we have $\phi\otimes \phi^{\Rightarrow x}=\phi$,
thus the axiom of idempotency holds.

Now, in order to prove that $(\Phi,D)$ is an information algebra, it suffices to check the axiom of combination, that is,
for $\phi,\psi\in \Phi$ and $x\in D$,
\begin{center}
$(\phi^{\Rightarrow x}\otimes \psi)^{\Rightarrow x}=\phi^{\Rightarrow x}\otimes \psi^{\Rightarrow x}$.
\end{center}
By the definition of focusing, it only need to show the case of $x=0$.\\
(\romannumeral 1) If $\phi\in[\frac{1}{2},1]$, then $(\phi^{\Rightarrow 0}\otimes \psi)^{\Rightarrow 0}
=\frac{1}{2}=\phi^{\Rightarrow 0}\otimes \psi^{\Rightarrow 0}$.\\
(\romannumeral 2) If $\phi\in[0,\frac{1}{2}],\psi\in[0,\frac{1}{2}]$, then $(\phi^{\Rightarrow 0}\otimes \psi)^{\Rightarrow 0}
=\phi\otimes\psi=\phi^{\Rightarrow 0}\otimes \psi^{\Rightarrow 0}$.\\
(\romannumeral 3) If $\phi\in[0,\frac{1}{2}],\psi\in[\frac{1}{2},1]$, then $(\phi^{\Rightarrow 0}\otimes \psi)^{\Rightarrow 0}
=\frac{1}{2}=\phi^{\Rightarrow 0}\otimes \psi^{\Rightarrow 0}$.

In summary, we have $(\phi^{\Rightarrow x}\otimes \psi)^{\Rightarrow x}=\phi^{\Rightarrow x}\otimes \psi^{\Rightarrow x}$.
Then $(\Phi,D)$ is a domain-free information algebra.}
\end{example}

If $(\Phi,D)$ is an information algebra, we write $\psi\leq \phi$,
means an information $\phi\in \Phi$ is more informative than another information
$\psi\in\Phi$, i.e., $\psi\otimes \phi=\phi$. The order relation $\leq$ is a partial order on an information algebra $(\Phi,D)$.
In this paper, the order relation $\leq$ induced by the operation of combination is a default order on an information algebra.

In Ref. 2, Kohlas gave a specific method to realize the transform between domain-free  information algebra and labeled information
algebra as follows.
In a labeled information algebra $(\Phi,D)$, we define for $\phi\in\Phi$ and $y\geq d(\phi)$,
$$\phi^{\uparrow y}=\phi\otimes e_y.$$
$\phi^{\uparrow y}$ is called the vacuous extension of $\phi$ to the domain $y$.
Now we consider a congruence relation $\sigma$:
\begin{center}
$\phi\equiv \psi (mod\ \  \sigma)$ if, and only if $\phi^{\uparrow x\vee y}=\psi^{\uparrow x\vee y}$,
\end{center}
where $x=d(\phi), y=d(\psi)$.
In the $(\Phi/\sigma,D)$, the two operations, combination and focusing, are defined as follows:
\begin{center}
Combination: $[\phi]_\sigma \otimes [\psi]_\sigma=[\phi\otimes \psi]_\sigma ;$\\
Focusing: $[\phi]_\sigma^{\Rightarrow x} =[(\phi^{\uparrow x\vee d(\phi)})^{\downarrow x}]_\sigma$.
\end{center}
Then $(\Phi/\sigma,D)$ is a domain-free information algebra, and
we say $(\Phi/\sigma,D)$ is the associated domain-free information algebra with $(\Phi,D)$.

Conversely, if $(\Phi,D)$ is a domain-free information algebra, let
\begin{center}
$\Psi=\{(\phi,x):\phi\in\Phi,\phi=\phi^{\Rightarrow x}\}$.
\end{center}
The three operations are defined on $\Psi$ as follows:

1. Labeling: For $(\phi,x)\in \Psi$ define $d(\phi,x)=x $;

2. Combination: For $(\phi,x),(\psi,y)\in \Psi$
define $(\phi,x)\otimes(\psi,y)=(\phi\otimes\psi,x\vee y)$;

3. Marginalization: For $(\phi,x)\in \Psi$ and $y\leq x$
define $(\phi,x)^{\downarrow y}=(\phi^{\Rightarrow y}, y)$.\\
Then $(\Psi, D)$ is a labeled information algebra, and is called the associated labeled information algebra with $(\Phi,D)$.

At the end of this section, we give some basic notions in lattice theory. Let $(L, \leq)$ be a partially ordered set.
$A$ is called a directed subset of $L$, if for all $a,b\in A$, there is a $c\in A$ such that $a,b\leq c$.
We write $\vee A$ for the least upper bound of $A$ in $L$ if it exists.
$L$ is called a sup-semilattice, if $a\vee b$ exists for all $a,b\in L$.
If every subset $A\subseteq L$ has a greatest lower bound or a least upper bound in $L$, we say $L$ is a complete lattice.

\begin{lemma}\label{lemma:3}\cite{Domain}
{\rm Let $L$ be a sup-semilattice with bottom element 0. Then $L$ is a complete lattice if, only if every directed subset
$A\subseteq L$ has the least upper bound $\vee A$.}
\end{lemma}

\begin{definition}\cite{Domain}
{\rm Let $L$ be a partially ordered set. For $a,b\in L$ we write $a\ll b$, and say $a$ way-below $b$ if,
for any directed set $X\subseteq L$, from $b\leq \vee X$ it follows that there is a $c\in X$
such that $a\leq c$. We call $a\in L$ a finite (compact) element, if $a\ll a$.}
\end{definition}

\begin{definition} \cite{scott}
{\rm For a complete lattice $L$, if for all $a\in L$, $a=\vee\{b:b\ll a\}$, we call $L$ a continuous lattice.
Moreover, if for all $a\in L,a=\vee \{b:b\ll b\leq a\}$, we call $L$ an algebraic lattice.}
\end{definition}

\section {Domain-free Continuous Information Algebras}\label{Semiring Induced Compact}

In this part we will give the concept of domain-free continuous information algebra,
and discuss the properties of continuous function spaces of domain-free continuous information algebras.
Information algebras mentioned in this section are all domain-free information algebras.

\subsection{Definitions}

Firstly, we give the following lemma which contains some simple and important results about the partially ordered relation
induced by the operation combination in information algebras.

\begin{lemma}\label{lemma:2}\cite{Information}
{\rm If $(\Phi,D)$ is an information algebra, then

1. $\phi^{\Rightarrow x}\leq \phi$.

2. $\phi\otimes\psi=\sup\{\phi,\psi\}$.

3. $\phi\leq \psi$ implies $\phi^{\Rightarrow x}\leq\psi^{\Rightarrow x}.$

4. $x\leq y$ implies $\phi^{\Rightarrow x}\leq\phi^{\Rightarrow y}.$}
\end{lemma}

In general, only ``finite" information can be treated in computers.
Therefore, for example in domain theory,
a structure that each information can be approximated by these ``finite" information has been proposed.
The concept of compact information algebra introduced by Kohlas \cite{Information} stems from the idea above.
Of course, the idea of approximation also prompt us to consider a type of information algebra that each element
$\phi$ in this system could be approximated with pieces of information which is ``relatively finite"
or \lq \lq way-below\rq \rq $\phi$. Then the notion of continuous information algebra is proposed next.

\begin{definition}\label{definition:2}
{\rm A system $(\Phi,\Gamma, D)$, where $(\Phi,D)$ is a domain-free information algebra, the lattice $D$ has a top element,
$\Gamma\subseteq\Phi$ is closed under combination and contains the empty information $e$,
satisfying the following axioms of convergence and density (resp. strong density),
is called a domain-free continuous (resp. s-continuous) information algebra.
$\Gamma$ is called a basis for the system $(\Phi, D)$.

1. Convergency: If $X\subseteq \Gamma$ is a directed set, then the supremum $\vee X$ exists.

2. Density(D1): For all $\phi\in \Phi$, $\phi=\vee \{\psi\in \Gamma: \psi\ll \phi\}.$

3. Strong density(SD1): For all $\phi\in \Phi$ and $x\in D$,
\begin{center}
$\phi^{\Rightarrow x}=\vee \{\psi\in \Gamma: \psi=\psi^{\Rightarrow x}\ll \phi\}.$
\end{center}

Moreover, if a domain-free continuous (resp. s-continuous) information algebra $(\Phi,\Gamma,D)$
satisfies the axiom of compactness,
then we call $(\Phi,\Gamma, D)$ a domain-free compact (resp. s-compact) information algebra.

4. Compactness: If $X\subseteq \Gamma$ is a directed set, and $\phi\in \Gamma$ such that $\phi\leq \vee X$ then there
exists a $\psi\in X$ such that $\phi\leq \psi$. }
\end{definition}

Sometime, for simplicity of expression, we directly say an information algebra is continuous or it has continuity
if it is a continuous information algebra. For the other concepts in the definition, there exist some similar forms of address.

\begin{lemma}\cite{Lecture}\label{lemma:1}
{\rm Let $(\Phi,\Gamma,D)$ be a compact information algebra, then the following holds:

1. $\psi\in \Gamma$ if, and only if $\psi\ll \psi$.

2. If $\psi\in \Gamma$, then $\psi\ll \phi$ if, and only if $\psi\leq \phi$.}
\end{lemma}

For an information algebra $(\Phi,D)$, we denote the set of all the finite elements of $\Phi$ by $\Phi_f$, i.e.,
$\Phi_f=\{\phi\in\Phi:\phi\ll\phi\}$.
If $(\Phi,\Gamma,D)$ is a compact information algebra, by Lemma \ref{lemma:1},
we have $\Gamma=\Phi_f$. Therefore, we always denote a (domain-free) compact information algebra by $(\Phi,\Phi_f,D)$.
In addition, by Lemma \ref{lemma:1} and the definition of way-below relation, we also can naturally get that,
an s-continuous(resp. continuous) information algebra $(\Phi,D)$ is s-compact(resp. compact)
if, and only if, the set $\Phi_f$ is a basis for the s-continuous(resp. continuous) information algebra $(\Phi,D)$.

A simple example of s-continuous information algebra but not s-compact information algebra is presented as follows.
\begin{example}\label{example:5}
{\rm Let an information algebra $(\Phi,D)$ with the operations combination and focusing be defined as in Example \ref{example:2}.

We are going to show that the equation
$\phi^{\Rightarrow x}=\vee \{\psi\in \Phi: \psi=\psi^{\Rightarrow x}\ll \phi\}$ is true for all $\phi\in\Phi$ and $x\in D$.
In fact, for the case of $x=0$,
if $\phi\in [0,\frac{1}{2}]$, then $\vee \{\psi\in [0,\frac{1}{2}]: \psi\ll\phi\}=\phi=\phi^{\Rightarrow 0}$;
otherwise, if $\phi\in [\frac{1}{2},1]$, then $\vee \{\psi\in [0,\frac{1}{2}]: \psi\ll\phi\}=\frac{1}{2}=\phi^{\Rightarrow 0}$.
Thus $\phi^{\Rightarrow 0}=\vee \{\psi\in [0,\frac{1}{2}]: \psi\ll \phi\}
=\vee \{\psi\in \Phi: \psi=\psi^{\Rightarrow 0}\ll \phi\}$ for all $\phi\in\Phi$.
For the case of $x=1$, the equation is clearly true.
Hence $(\Phi,D)$ is s-continuous and $\Phi$ is a basis. But it is not s-compact, because $\Phi_f=\{0\}$.}
\end{example}

\begin{proposition}\label{proposition:5}
If $(\Phi,\Gamma,D)$ is an s-continuous information algebra, then
$\{\psi\in \Gamma: \psi^{\Rightarrow x}=\psi\ll \phi\}$ is directed for all $\phi\in \Phi$ and $x\in D$.

Similarly, if $(\Phi,\Gamma,D)$ is a continuous information algebra, then
$\{\psi\in \Gamma: \psi\ll \phi\}$ is directed for all $\phi\in \Phi$.
\end{proposition}
\noindent {\bf Proof.} We only prove the case of an s-continuous information algebras.

Let $\psi_1,\psi_2\in\{\psi\in \Gamma:\psi^{\Rightarrow x}=\psi\ll\phi\}$.
Since $\Gamma$ is closed under combination, we have $\eta\in \Gamma$ if $\eta=\psi_1\otimes\psi_2$.
By the monotonicity of the operation focusing, we have
$\psi_i=\psi_i^{\Rightarrow x}\leq \eta^{\Rightarrow x}(i=1,2)$. So $\eta\leq \psi^{\Rightarrow x}\leq \eta$,
that is, $\eta=\eta^{\Rightarrow x}$.
Next we show that $\eta\ll\phi$. If $X\subseteq \Phi$ is directed and $\phi\leq \vee X$, by $\psi_1,\psi_2\ll\phi$,
there exist $\phi_1,\phi_2\in X$ such that $\psi_1\leq\phi_1$ and $\psi_2\leq\phi_2$.
Since $X$ is directed, there is a $\phi_3\in X$ such that $\phi_1,\phi_2\leq\phi_3$.
Then $\psi_1,\psi_2\leq\phi_3$. We obtain that $\eta\leq\phi_3$ and thus $\eta\ll\phi$.
By what we have proved, we have $\eta\in \{\psi\in \Gamma: \psi=\psi^{\Rightarrow x}\ll\phi\}$.
Hence $\{\psi\in \Gamma: \psi=\psi^{\Rightarrow x}\ll\phi\}$ is a directed set.
\qed

\begin{proposition}\label{proposition:1}
{\rm Let $(\Phi,D)$ be an information algebra. The lattice $D$ has a top element.
Then $(\Phi,D)$ is continuous (resp. s-continuous) if, and only if
there exists a set $\Upsilon\subseteq \Phi$ such that $\Upsilon$ contains the empty information $e$ and satisfies the two
conditions of convergency and density(D2) (resp. strong density(SD2)):

1. Convergency: If $X\subseteq \Upsilon$ is a directed set, then the supremum $\vee X$ exists.

2. Density(D2): For all $\phi\in \Phi$, $\{\psi\in \Upsilon: \psi\ll \phi\}$ is directed and
\begin{center}
$\phi=\vee \{\psi\in \Upsilon: \psi\ll \phi\}.$
\end{center}

3. Strong density(SD2): For all $\phi\in \Phi$ and $x\in D$,
$\{\psi\in \Upsilon: \psi=\psi^{\Rightarrow x}\ll \phi\}$ is directed and
\begin{center}
$\phi^{\Rightarrow x}=\vee \{\psi\in \Upsilon: \psi=\psi^{\Rightarrow x}\ll \phi\}.$
\end{center}}
\end{proposition}
\noindent {\bf Proof.}
By Proposition \ref{proposition:5}, the necessary condition is clear.

Conversely, assume that $\Upsilon$ is a set which satisfies the assumptions. We need to find a basis $\Gamma$ for $(\Phi,D)$.

We show that $(\Phi, \leq)$ is a complete lattice
if $\Upsilon$ satisfies the two conditions of convergency and density(D2).
Firstly, $\{\psi\in \Upsilon: \psi\ll\phi,\phi\in X\}$ is also directed if $X\subseteq \Phi$ is directed.
Assume that $\psi_1,\psi_2\in \{\psi\in \Upsilon: \psi\ll\phi,\phi\in X\}$,
then there exist $\phi_1,\phi_2\in X$ such that $\psi_1\ll\phi_1,\psi_2\ll\phi_2$.
So there is a $\phi\in X$ such that $\psi_1,\psi_2\ll\phi$ because $X$ is directed.
Since $\phi=\vee \{\psi\in \Upsilon: \psi\ll \phi\}$,
by the definition of way-below relation and the directness of set $\{\psi\in \Upsilon: \psi\ll \phi\}$,
there exists a $\eta\in \Upsilon $ such that $\psi_1,\psi_2\leq \eta\ll \phi$.
This proves that $\{\psi\in \Upsilon: \psi\ll\phi,\phi\in X\}$ is also directed. By the axiom of convergency,
$\vee\{\psi\in \Upsilon: \psi\ll\phi,\phi\in X\}$ exists. Obviously, $\vee X= \vee\{\psi\in \Upsilon: \psi\ll\phi,\phi\in X\}$.
By Lemma \ref{lemma:2}(2), $(\Phi,D)$ is a sup-semilattice. Since empty information $e\in \Phi$ is the bottom element,
we obtain that $(\Phi,\leq)$ is a complete lattice by Lemma \ref{lemma:3}.

Clearly the condition of strong density(SD2) is stronger than density(D2), then
$(\Phi, \leq)$ is also a complete lattice if $\Upsilon$ satisfies the convergency and strong density(SD2).

Now we claim that $\Gamma=\Phi$ which is closed under combination is a basis for the system.
Firstly, since $(\Phi,\leq)$ is a complete lattice,
the condition of convergency of $\Gamma$ holds, and $\vee \{\psi\in \Gamma: \psi\ll \phi\}$,
$\vee \{\psi\in \Gamma: \psi=\psi^{\Rightarrow x}\ll \phi\}$ exist.

If the axiom of density(D2) holds, then
$\phi=\vee \{\psi\in \Upsilon: \psi\ll \phi\}\leq \vee \{\psi\in \Gamma: \psi\ll \phi\}\leq \phi$.
We conclude that $\vee \{\psi\in \Gamma: \psi\ll \phi\}=\phi$.
Hence $\Gamma$ is a basis for the continuous information algebra $(\Phi,D)$.

Similarly, if the axiom of strong density(SD2) holds,
then $\phi^{\Rightarrow x}
=\vee \{\psi\in \Upsilon: \psi=\psi^{\Rightarrow x}\ll \phi\}
\leq \vee \{\psi\in \Gamma: \psi=\psi^{\Rightarrow x}\ll \phi\}\leq \phi^{\Rightarrow x}$.
We obtain that $\vee \{\psi\in \Gamma: \psi=\psi^{\Rightarrow x}\ll \phi\}=\phi^{\Rightarrow x}$.
Hence $\Gamma$ is a basis for the s-continuous information algebra $(\Phi,D)$.
\qed

By Proposition \ref{proposition:1}, we give an equivalent definition of continuous information algebra.
And we know that the subset of a basis $\Gamma$ consisting of all the elements which approximate to
an element $\phi$ in a continuous information algebra is directed.
It is convenient for discussing the problem about continuous functions which is defined in the next subsection.
But, it is worth noting that the set $\Upsilon$ in the Proposition \ref{proposition:1},
satisfying the convergency and density(D2), is not necessarily closed under combination,
although we can find out a basis $\Gamma$ which is closed under combination for the system.
An simple example rooted in lattice theory is shown as follows.

\begin{example}\label{example:1}
{\rm Let ${\bf X}$ be an infinite set, $\Phi={\cal P}(X)$ and $D=\{1\}$.
We define the operations combination and focusing as follows:
\begin{center}
Combination: $A \otimes B=A\cup B, \forall A, B\in \Phi$;\\
Focusing: $A^{\Rightarrow 1}=A, \forall A\in \Phi.$
\end{center}

The system $(\Phi,D)$ is an information algebra.
Let $\Gamma$ be the set of all the finite subsets of $X$. It is clear that $C\ll C$ if $C\in \Gamma$.
For each set $A\subseteq X$, it can be represented by the combination of the directed family of all finite subsets of $A$,
i.e., $A=\vee \{B\in \Gamma: B\ll A\}$ and $\{B\in \Gamma: B\ll A\}$ is directed.
Suppose that $Y\subset X$ is an infinite proper subset. We write $\Gamma^*=\Gamma\cup\{Y\}$.
Obviously, for all $A\in \Phi$, $A=\vee \{B\in \Gamma^*: B\ll A\}$ and $\{B\in \Gamma^*: B\ll A\}$ is still directed.
While $\Gamma^*$ is not closed under combination. For example, we have $\{x\}, Y\in \Gamma^*$ where $x\in X-Y$,
but $\{x\}\otimes Y\not\in \Gamma^*$.}
\end{example}

According to the proof in Proposition \ref{proposition:1}, the following theorem can be drawn immediately.
It gives a compact expression for these notions defined in Definition \ref{definition:2}.

\begin{theorem}\label{theorem:8}
{\rm Let $(\Phi, D)$ be an information algebra.

1. $(\Phi, D)$ is continuous (resp. compact) if, and only if $(\Phi,\leq)$ is a continuous lattice(resp. algebraic lattice).

2. $(\Phi, D)$ is s-continuous (resp. s-compact) if, and only if $(\Phi,\leq)$ is a complete lattice and
for all $\phi\in \Phi$, $x\in D$,
\begin{center}
$\phi^{\Rightarrow x}=\vee\{\psi\in \Phi: \psi=\psi^{\Rightarrow x}\ll \phi\}.$

(resp. $\phi^{\Rightarrow x}=\vee\{\psi\in \Phi_f: \psi=\psi^{\Rightarrow x}\leq \phi\}.$)
\end{center}
}
\end{theorem}
\noindent {\bf Proof.} Here we take the proof of the equivalent definition of s-compact information algebra as instance.
If $(\Phi, D)$ is s-compact, then $(\Phi,\leq)$ is a complete lattice
which can be concluded from the proof of Proposition \ref{proposition:1}.

Conversely, $\Phi_f$ satisfies the axiom of convergency if $(\Phi,\leq)$ is a complete lattice.
The compactness of $\Phi_f$ is directly obtained from the definition of way-below relation and the condition of
$\phi\ll \phi$ if $\phi\in\Phi_f$. If $\phi,\psi\in\Phi_f$, we claim that $\phi\otimes\psi\in\Phi_f$.
In fact, if $X\subseteq \Phi$ is a directed subset and $\phi\otimes\psi\leq \vee X$, by $\phi\ll\phi$ and $\psi\ll\psi$,
there exists a $\eta\in X$ such that $\phi,\psi\le\eta$. So $\phi\otimes\psi\leq\eta$. Hence $\phi\otimes\psi\ll
\phi\otimes\psi$, i.e., $\phi\otimes\psi\in\Phi_f$. We obtain that $\Phi_f$ is a basis for the s-compact information
algebra $(\Phi, D)$.
\qed

\subsection{Continuous Functions}
In Ref. 2, the notion of continuous function between two s-compact information algebras has been introduced.
In this subsection, the definition of continuous functions is extended to be a more general situation.
We focus on the problem that whether a function space which is the set consisting of all continuous functions between
two continuous information algebras can still form a continuous information algebra.
In fact, the discussions about function spaces have appeared in various areas of mathematics and computer science.
For example, in category theory, if the problem here is true, it is benefit to create a Cartesian closed category.

\begin{definition}\label{definition:3}
{\rm A mapping $f: \Phi\rightarrow \Psi$ from a continuous information algebra
$(\Phi,D)$ into another continuous information algebra $(\Psi,E)$ is called continuous,
if for every directed subset $X \subseteq \Phi$, $f(\vee X) = \vee f(X)$.}
\end{definition}

The form of the definition is consistent with the definition of Scott continuous functions in domain theory.
A function $f$ between two complete lattices $S,T$ is called Scott continuous,\cite{scott} if it preserves
the supremum of all directed sets, i.e., $f(\vee X) = \vee f(X)$ for all directed subset $X\subseteq S$.
Let $[\Phi\rightarrow \Psi]_c$ denote the set of all continuous mappings from $(\Phi,D)$ into $(\Psi,E)$.
We define the two operations of combination and focusing on the system
 $([\Phi\rightarrow \Psi]_c, D\times E)$ as follows \cite{Information}:

1. Combination: for $f,g \in [\Phi\rightarrow \Psi]_c$ define $f\otimes g$ by
$$(f\otimes g)(\phi)=f(\phi)\otimes g(\phi).$$

2. Focusing: for $f\in [\Phi\rightarrow \Psi]_c, (x,y)\in D\times E$ define $f^{\Rightarrow(x,y)}$ by
$$f^{\Rightarrow(x,y)}(\phi)=(f(\phi^{\Rightarrow x}))^{\Rightarrow y}.$$

\begin{lemma}\label{lemma:8}\cite{Domain}
{\rm If $S,T$ are two continuous lattices, then $[S\rightarrow T]_c$ is a continuous lattice
with respect to the pointwise partial order.}
\end{lemma}

\begin{theorem}\label{theorem:6}
{\rm If $(\Phi,D)$ is continuous, then it is s-continuous if, and only if, for all directed subset $X$ of $\Phi$ and $x\in D$,
\begin{center}
 $(\vee X)^{\Rightarrow x}=\mathop\bigvee\limits_{\phi\in X} \phi^{\Rightarrow x}.$
\end{center}  }
\end{theorem}
\noindent {\bf Proof.}
Let $(\Phi,D)$ be s-continuous. For all directed subset $X$ of $\Phi$ and $x\in D$,
by Lemma \ref{lemma:2}, we have
$(\vee X)^{\Rightarrow x}\geq{\mathop\bigvee\limits_{\phi\in X}} \phi^{\Rightarrow x}.$
On the other hand,
by density, $(\vee X)^{\Rightarrow x}=\vee\{\psi\in \Gamma: \psi=\psi^{\Rightarrow x}\ll \vee X\}$,
where $\Gamma$ is a basis for $(\Phi,D)$. If $\psi\ll \vee X$,
then there exists a $\phi\in X$ such that $\psi\leq\phi$. Then $\psi=\psi^{\Rightarrow x}\leq \phi^{\Rightarrow x}$.
Hence $(\vee X)^{\Rightarrow x}\leq \mathop\bigvee\limits_{\phi\in X} \phi^{\Rightarrow x}$.
So we get that $(\vee X)^{\Rightarrow x}=\mathop\bigvee\limits_{\phi\in X} \phi^{\Rightarrow x}.$

Conversely, if $(\Phi,D)$ is continuous, then $\phi=\vee \{\psi\in \Phi: \psi\ll \phi\}$
and $\{\psi\in \Phi: \psi\ll \phi\}$ is directed for all $\phi\in\Phi$.
Thus, by assumption, we have $\phi^{\Rightarrow x}=\vee \{\psi^{\Rightarrow x}:\psi\in \Phi, \psi\ll \phi\}$.
Meanwhile,
$\{\psi^{\Rightarrow x}:\psi\in \Phi, \psi\ll \phi\} \subseteq \{\varphi\in\Phi:\varphi=\varphi^{\Rightarrow x}\ll\phi\}$,
because $\psi^{\Rightarrow x}=(\psi^{\Rightarrow x})^{\Rightarrow x}$ for all $\psi\in\Phi$.
Then $\phi^{\Rightarrow x}\leq \vee \{\varphi\in\Phi:\varphi=\varphi^{\Rightarrow x}\ll\phi\}\leq \phi^{\Rightarrow x}$
by Lemma \ref{lemma:2}, that is,
$\phi^{\Rightarrow x}= \vee \{\varphi\in\Phi:\varphi=\varphi^{\Rightarrow x}\ll\phi\}$.
This shows that $(\Phi,D)$ is s-continuous.
\qed

\begin{proposition}\label{proposition:3}
{\rm If $(\Phi,D)$ and $(\Psi,E)$ are all continuous,
$f,g \in [\Phi\rightarrow \Psi]_c$ and $(x,y)\in D\times E$, then $f\otimes g\in [\Phi\rightarrow \Psi]_c$.

Moreover, if $(\Phi,D)$ and $(\Psi,E)$ are s-continuous, then
$f^{\Rightarrow(x,y)}\in [\Phi\rightarrow \Psi]_c$.}
\end{proposition}
\noindent {\bf Proof.} Let $\{\phi_i:i\in I\}$ be a directed subset of $\Phi$. We have
\begin{eqnarray*}
\begin{array}{lll}
&&(f\otimes g)(\mathop\bigvee\limits_{i\in I}\phi_i)
=f(\mathop\bigvee\limits_{i\in I}\phi_i) \otimes g(\mathop\bigvee\limits_{i\in I}\phi_i)\\
&=&[\mathop\bigvee\limits_{i\in I} f(\phi_i)] \otimes [\mathop\bigvee\limits_{i\in I} g(\phi_i)]
=\mathop\bigvee\limits_{i\in I} [f(\phi_i) \otimes g(\phi_i)]\\
&=&\mathop\bigvee\limits_{i\in I} (f \otimes g)(\phi_i),
\end{array}
\end{eqnarray*}
then $f\otimes g\in [\Phi\rightarrow \Psi]_c$.

Suppose $X\subseteq\Phi$ is a directed set.
Since $f$ is order-preserving, we have $\{f(\phi^{\Rightarrow x}):\phi\in X\}$ is also directed.
By Theorem \ref{theorem:6}, we have
\begin{eqnarray*}
\begin{array}{lll}
 &&f^{\Rightarrow(x,y)}(\vee X)
=(f((\vee X)^{\Rightarrow x}))^{\Rightarrow y} \\
&=&(f(\mathop\bigvee\limits_{\phi\in X} \phi^{\Rightarrow x}))^{\Rightarrow y}
=(\mathop\bigvee\limits_{\phi\in X} f(\phi^{\Rightarrow x}))^{\Rightarrow y}\\
&=&\mathop\bigvee\limits_{\phi\in X} (f(\phi^{\Rightarrow x}))^{\Rightarrow y}
=\mathop\bigvee\limits_{\phi\in X} f^{\Rightarrow (x,y)}(\phi).
\end{array}
\end{eqnarray*}
Then $f^{\Rightarrow(x,y)}\in [\Phi\rightarrow \Psi]_c$.
\qed

\begin{lemma}\label{lemma:9}
{\rm If $(\Phi,D)$ and $(\Psi,E)$ are s-continuous, then $([\Phi\rightarrow \Psi]_c, D\times E)$ is continuous.}
\end{lemma}
\noindent {\bf Proof.}
Firstly, we claim that $([\Phi\rightarrow \Psi]_c, D\times E)$ is a domain-free information algebra.

1. Semigroup: If $e_2$ is a neutral element of $(\Psi,E)$, let $h: \Phi\rightarrow\Psi$ be defined as
$h(\phi)=e_2,\forall \phi\in\Phi$. Then $h$ is a neutral element of $[\Phi\rightarrow \Psi]_c$.
It's obvious that $[\Phi\rightarrow \Psi]_c$ is associative and commutative under combination.

2. Idempotency: For $f\in [\Phi\rightarrow \Psi]_c$ and $(x,y)\in D\times E$,
because $f$ is order-preserving, we obtain that
$(f(\phi^{\Rightarrow x}))^{\Rightarrow y}\leq f(\phi^{\Rightarrow x})\leq f(\phi)$ for all $\phi\in\Phi$.
Then $(f\otimes f^{\Rightarrow(x,y)})(\phi)
 =f(\phi)\otimes f^{\Rightarrow(x,y)}(\phi)
 =f(\phi)\otimes (f(\phi^{\Rightarrow x}))^{\Rightarrow y}
 =f(\phi)$.
Hence $f\otimes f^{\Rightarrow(x,y)}=f$.

The axioms of transitivity, combination and support are easy to show.
Hence $([\Phi\rightarrow \Psi]_c, D\times E)$ is a domain-free information algebra.

Since $(\Phi,\leq)$ and $(\Psi,\leq)$ are two continuous lattices, by Lemma \ref{lemma:8},
we have $([\Phi\rightarrow\Psi]_c, \leq_p)$ is also continuous,
where $\leq_p$ means the pointwise order on $[\Phi\rightarrow\Psi]_c$.
For all $f,g\in [\Phi\rightarrow\Psi]_c$, we have
\begin{eqnarray*}
\begin{array}{lll}
 f\leq_p g &\Leftrightarrow& \forall \phi\in\Phi, f(\phi)\leq g(\phi)\\
&\Leftrightarrow & \forall \phi\in\Phi, f(\phi)\otimes g(\phi)=g(\phi),{\rm that\ \  is}, (f\otimes g)(\phi)=g(\phi)\\
& \Leftrightarrow & f\otimes g=g, {\rm that\ \  is}, f\leq g.
\end{array}
\end{eqnarray*}
Then $([\Phi\rightarrow\Psi]_c, \leq)$ is a continuous lattice.
This shows that $([\Phi\rightarrow \Psi]_c, D\times E)$ is a continuous information algebra by Theorem \ref{theorem:8}.
\qed

\begin{theorem}\label{theorem:7}
{\rm If $(\Phi,D)$ and $(\Psi,E)$ are s-continuous,
then $([\Phi\rightarrow \Psi]_c,D\times E)$ is also s-continuous.}
\end{theorem}
\noindent {\bf Proof.}
We have known that $([\Phi\rightarrow \Psi]_c, D\times E)$ is a continuous information algebra by Lemma \ref{lemma:9}.

Let $\{f_i:i\in I\}\subseteq[\Phi\rightarrow \Psi]_c$ be directed.
For all $\phi\in\Phi, (x,y)\in D\times E$, we have
\begin{eqnarray*}
\begin{array}{lll}
&&(\mathop\bigvee\limits_{i\in I} f_i)^{\Rightarrow (x,y)}(\phi)
=[(\mathop\bigvee\limits_{i\in I} f_i)(\phi^{\Rightarrow x})]^{\Rightarrow y}\\
&=&[\mathop\bigvee\limits_{i\in I} f_i(\phi^{\Rightarrow x})]^{\Rightarrow y}
=\mathop\bigvee\limits_{i\in I} [f_i(\phi^{\Rightarrow x})]^{\Rightarrow y}\\
&=&\mathop\bigvee\limits_{i\in I} f_i^{\Rightarrow (x,y)}(\phi)
=(\mathop\bigvee\limits_{i\in I} f_i^{\Rightarrow (x,y)})(\phi).
\end{array}
\end{eqnarray*}
Thus $(\mathop\bigvee\limits_{i\in I} f_i)^{\Rightarrow (x,y)}=\mathop\bigvee\limits_{i\in I} f_i^{\Rightarrow (x,y)}$.
By Theorem \ref{theorem:6}, $([\Phi\rightarrow \Psi]_c,D\times E)$ is an s-continuous information algebra.
\qed

\section {Labeled Continuous Information Algebra}

In this section, we will consider a type of continuous information algebras with the operation of labeling,
and study the relationship between labeled continuous information algebras and domain-free continuous information algebras.

In Ref. 3, a notion of labeled compact information algebra has been presented.
It also has given a counterexample to claim that labeled compact information algebra
does not necessarily lead to domain-free compact information algebra.
We will present an improved definition of labeled compact information algebra such that
the associated domain-free information algebras with labeled compact information algebras are compact.

We firstly introduce these definitions about continuity and compactness in labeled information algebras.
The difference of continuity between labeled information algebras and domain-free information algebras
is that, in a labeled continuous information algebra,
every element $\phi\in \Phi_x$ can be approximated by some elements with domain $x$ which are way-below $\phi$.
Therefore, labeled continuous information algebras are a kind of information algebras which have continuity in local domain.

\begin{definition}
{\rm A system $(\Phi,\{\Gamma_x\}_{x\in D}, D)$, where $(\Phi, D)$ is a labeled information algebra and
the lattice $D$ has a top element $\top$, is called a labeled continuous (resp. s-continuous) information algebra, if,
for all $x\in D$, $\Gamma_x\subseteq \Phi_x$ is closed under the combination,
contains a neutral element $e_x$ and satisfies the following axioms of convergency and density (resp. strong density):

1. Convergency: If $X\subseteq \Gamma_x$ is a directed set, then the supremum $\vee X$ exists and $\vee X\in \Phi_x$.

2. Density: For all $\phi\in \Phi_x$, $\phi=\vee \{\psi\in \Gamma_x: \psi\ll_x \phi\}$,
where $\psi\ll_x \phi$ means $\psi\ll\phi$ in $\Phi_x$.

3. Strong density: For all $\phi\in \Phi_x,
\phi=\vee \{ {\psi ^{ \downarrow x}} \in {\Gamma_x}:{\psi^{\downarrow x}} \otimes e_\top \in {\Gamma_\top},
 {\psi^{\downarrow x}}{ \ll _x}\phi\}.$

Moreover, if a labeled continuous (resp. s-continuous) information algebra
$(\Phi,\{\Gamma_x\}_{x\in D}, D)$ satisfies the axiom of
compactness, we call it a labeled compact (resp. s-compact) information algebra.

4. Compactness: If $X\subseteq \Gamma_x$ is a directed set, and $\phi\in \Gamma_x$ such that $\phi\leq \vee X$ then there
exists a $\psi\in X$ such that $\phi\leq \psi$.}
\end{definition}

\begin{remark}\label{remark:1}
It is clear that the condition of strong density is stronger than the condition of density.
Then s-continuous information algebras must be continuous.
By using the same method as in the proof of Proposition \ref{proposition:1}, we can also get that $(\Phi_x, \leq)$
is a complete lattice for all $x\in D$, if $(\Phi,D)$ is continuous.
\end{remark}

\begin{lemma}\label{lemma:4}
{\rm Let $(\Phi,\{\Gamma_x\}_{x\in D}, D)$ be a labeled compact information algebra. The following holds:

1.$\psi\in \Gamma_x$ if, and only if $\psi\ll_x \psi$.(see Ref. 3)

2. If $\psi\in \Gamma_x$ and $\phi\in \Phi_x$, then $\psi\ll_x \phi$ if, and only if $\psi\leq \phi$.}
\end{lemma}
The above lemma follows easily from the definitions.
By Lemma \ref{lemma:4}, we can denote a labeled compact information algebra by $(\Phi,\{\Phi_{f,x}\}_{x\in D},D)$
or $(\Phi,\Phi_{f,x},D)$,
where $\Phi_{f,x}=\{\psi\in \Phi_x:\psi\ll_x\psi\}$.

Let us see an example of a labeled compact information algebra.
\begin{example}\label{example:6}
In Example \ref{example:3}, we consider a labeled information algebra $({\cal F},{\cal P}(E))$.
Here, for a soft set $(F,A)$, we assume that set $A\subseteq E$ is finite.
This assumption is consistent with the application in real-life.
The partial order $\leq$ induced by the operation $\sqcap$ is defined as: $(F,A)\leq (G,B)$ if

(\romannumeral 1) $A \subseteq B$, and

(\romannumeral 2) $\forall e\in A$, $G(e)$ is a subset of $F(e)$.

Firstly, $({\cal F}_A,\leq)$ is a complete lattice. In fact, for a set $\{(F_i,A):i\in I\}\subseteq {\cal F}_A$, we have
$\vee \{(F_i,A):i\in I\}=(F,A)$, where $(F,A)$ is defined as $F(e)=\bigcap\limits_{i\in I} F_i(e)$ for all $e\in A$.

Secondly, we demonstrate that $(F,A)\ll_A (F,A)$, i.e.,$(F,A)\in {\cal F}_{f,A}$
if, and only if, $\forall e\in A$, $U-F(e)$ is a finite subset of $U$.

(1) Let $\{(G_i,A):i\in I\}$ be a directed set and $(F,A)\leq \bigvee\limits_{i\in I} (G_i,A)$.
We write $\bigvee\limits_{i\in I} (G_i,A)=(G,A)$.
$\forall e\in A$, we have $G(e)=\bigcap\limits_{i\in I} G_i(e)\subseteq F(e)$.
Then $U-F(e)\subseteq U-G(e)=\bigcup\limits_{i\in I} (U-G_i(e))$.
Since $U-F(e)$ is a finite set and $\{(G_i,A):i\in I\}$ is directed,
there exists a $i^{(e)}\in I$ such that $U-F(e)\subseteq U-G_{i^{(e)}}(e)$.
Then $G_{i^{(e)}}(e)\subseteq F(e)$.
Since $A$ is a finite set and $\{(G_i,A):i\in I\}$ is directed again, there exists a $j\in I$ such that $G_j(e)\subseteq F(e)$
for all $e\in A$, that is, $(F,A)\leq (G_j,A)$. This proves $(F,A)\ll_A (F,A)$.

(2) For all $e\in A$, $U-F(e)$ can be represented as the supremum of $\{B_i:i\in I\}$  which is a directed family of
all the finite subsets of $U-F(e)$, i.e., $U-F(e)=\bigcup\limits_{i\in I} B_i$. We define a family of soft sets $(H_i,A)$ as
follows:
$$H_i(\varepsilon)=\left\{
          \begin {array}{ll}
          U-B_i,&{\mbox{if}} \ \ \varepsilon=e;\\
          F(e),&{\mbox{otherwise}}.
         \end{array}
        \right.$$
With respect to the order relation $\leq$, we have $\{(H_i,A):i\in I\}$ is a directed subsets of ${\cal F}_A$
and $(F,A)= \bigvee\limits_{i\in I} (H_i,A)$.
Since $(F,A)\ll_A (F,A)$, there exists a $k\in I$ such that $(F,A)\leq (H_k,A)$.
Hence $U-F(e)\subseteq U-H_k(e)=B_k$. Thus $U-F(e)$ is a finite subset of $U$. This proves what we have stated.

At last, we need to show the following equation holds for all $A\subseteq E$,
$$(F,A)=\vee\{(G,A)\in {\cal F}_{f,A}: (G,A)\leq (F,A)\}.$$

For all $e\in A$, $U-F(e)$ can be represented as the supremum of $\{B_i:i\in I^{(e)}\}$  which is a directed family of
all the finite subsets of $U-F(e)$, i.e., $U-F(e)=\bigcup\limits_{i\in I^{(e)}} B_i$. We define a family of soft sets $(F_i,A)$ as
follows:
$$F_i(\varepsilon)=\left\{
          \begin {array}{ll}
          U-B_i,&{\mbox{if}} \ \ \varepsilon=e;\\
          U,&{\mbox{otherwise}}.
         \end{array}
        \right.$$
Therefore, according to the above observation, we have $\{(F_i,A):i\in I^{(e)},e\in A\}\subseteq {\cal F}_{f,A}$.
We write $\bigvee\limits_{i\in I^{(e)},e\in A} (F_i,A)=(H,A)$.
For all $d\in A$, $H(d)=\bigcap\limits_{i\in I^{(e)},e\in A} F_i(d)=\bigcap\limits_{i\in I^{(d)}} F_i(d)
=\bigcap\limits_{i\in I^{(d)}} (U-B_i)=F(d)$. Therefore
$(F,A)=(H,A)=\bigvee\limits_{i\in I^{(e)},e\in A} (F_i,A)\leq \vee\{(G,A)\in {\cal F}_{f,A}: (G,A)\leq (F,A)\}\leq (F,A)$.
Hence $(F,A)=\vee\{(G,A)\in {\cal F}_{f,A}: (G,A)\leq (F,A)\}.$

Now we can say that $({\cal F},{\cal P}(E))$ is a labeled compact information algebra.
\end{example}

\begin{lemma}\cite{Lecture}\label{lemma:5}
{\rm Let $(\Phi,D)$ be a domain-free information algebra.
$(\Psi,D)$ is the labeled information algebra associated with $(\Phi,D)$.
For any set $X\subseteq \Psi_x$, if $\mathop\bigvee\limits_{(\psi,x)\in X} \psi$ exists, then
\begin{center}
$\mathop\bigvee\limits_{(\psi,x)\in X} (\psi,x)=(\mathop\bigvee\limits_{(\psi,x)\in X} \psi,x).$
\end{center}}
\end{lemma}

\begin{lemma}\label{lemma:6}
{\rm Let $(\Psi,D)$ be the labeled information algebra associated with a domain-free continuous information algebra $(\Phi,D)$.
If $\psi=\psi^{\Rightarrow x}$ and $\phi=\phi^{\Rightarrow x}$,
then $\psi\ll\phi$ implies $(\psi,x)\ll_x (\phi,x)$.

Furthermore, let $(\Phi,D)$ be a domain-free s-continuous information algebra, if
$\psi=\psi^{\Rightarrow x}$ and $\phi=\phi^{\Rightarrow x}$,
then $\psi\ll\phi$ if, and only if $(\psi,x)\ll_x (\phi,x)$.}
\end{lemma}
\noindent {\bf Proof.}
Assume that $(\Phi,D)$ is a domain-free continuous information algebra and
$\{(\varphi_j,x)\}_{j\in J}$ is a directed subset of $\Psi_x$ such that
$(\phi,x)\leq \mathop\bigvee\limits_{j\in J}(\varphi_j,x)$, by Lemma \ref{lemma:5},
we have $(\phi,x)\leq (\mathop\bigvee\limits_{j\in J} \varphi_j,x)$.
Thus $\phi\leq \mathop\bigvee\limits_{j\in J} \varphi_j$. Since $\psi\ll\phi$, there is a $j\in J$ such that $\psi\leq \varphi_j$.
Then $(\psi,x)\leq (\varphi_j,x)$. We have $(\psi,x)\ll_x (\phi,x)$.

On the other hand, if $(\psi,x)\ll_x (\phi,x)$ and $\{\varphi_j\}_{j\in J}$ is a directed subset of $\Phi$ such that
$\phi\leq \mathop\bigvee\limits_{j\in J} \varphi_j$, by Theorem \ref{theorem:5},
we have $\phi\leq \mathop\bigvee\limits_{j\in J} \varphi_j^{\Rightarrow x}$.
Therefore, $(\phi,x)\leq (\mathop\bigvee\limits_{j\in J} \varphi_j^{\Rightarrow x},x)
=\mathop\bigvee\limits_{j\in J}(\varphi_j^{\Rightarrow x},x)$.
By the definition of way-below relation, there exists a $j\in J$ such that $(\psi,x)\leq (\varphi_j^{\Rightarrow x},x)$.
We have $\psi\leq \varphi_j^{\Rightarrow x}\leq \varphi_j$. Hence $\psi\ll\phi$.
\qed

\begin{theorem}\label{theorem:2}
{\rm If $(\Phi,D)$ is a domain-free s-continuous information algebra, then its associated labeled information algebra
is s-continuous too.}
\end{theorem}
\noindent {\bf Proof.} Let $(\Psi,D)$ with $\Psi=\{(\phi,x):\phi\in\Phi,\phi=\phi^{\Rightarrow x}\}$
be the labeled information algebra associated with $(\Phi,D)$.
We define $\Gamma_x=\{(\phi,x):\phi\in \Gamma,\phi=\phi^{\Rightarrow x}\}$, where $\Gamma$ is the basis for $(\Phi,D)$.

1. If $(\phi_1,x), (\phi_2,x)\in \Gamma_x$, we have $\phi_1\otimes \phi_2\in \Gamma$ and $(\phi_1\otimes \phi_2)^{\Rightarrow x}
=(\phi_1^{\Rightarrow x}\otimes \phi_2)^{\Rightarrow x}=\phi_1^{\Rightarrow x}\otimes \phi_2^{\Rightarrow x}
=\phi_1\otimes \phi_2$, then $(\phi_1\otimes \phi_2, x)\in \Gamma_x$. It is clear that $(e,x)\in \Gamma_x$ and $(e,x)$
is a neutral element.

2. Let $X\subseteq \Gamma_x$ be a directed set. Then $\{\phi:(\phi,x)\in X\}\subseteq \Gamma$ is also directed,
and $\mathop\vee \{\phi:(\phi,x)\in X\}$ exists. By Lemma \ref{lemma:5},
$\vee X=(\mathop\bigvee\limits_{(\phi,x)\in X} \phi,x)\in \Psi_x$.

3. For all $(\phi,x)\in \Psi_x$, by Lemma \ref{lemma:5} and Lemma \ref{lemma:6}, we have
\begin{eqnarray*}
\begin{array}{lll}
&&\vee \{(\psi,z)^{\downarrow x}\in \Gamma_x: (\psi,z)^{\downarrow x}\otimes e_\top\in \Gamma_\top,
      (\psi,z)^{\downarrow x}\ll_x (\phi,x)\}\\
&=&\vee \{(\psi^{\Rightarrow x},x)\in \Gamma_x: (\psi^{\Rightarrow x},x)\otimes e_\top\in \Gamma_\top,
          (\psi^{\Rightarrow x},x)\ll_x (\phi,x)\}\\
&=& \vee \{(\psi^{\Rightarrow x},x): \psi^{\Rightarrow x}\in \Gamma, \psi^{\Rightarrow x}\ll \phi\}\\
&=& (\vee \{\psi^{\Rightarrow x}\in \Gamma: \psi^{\Rightarrow x}\ll\phi\},x)\\
&=& (\vee \{\psi\in \Gamma: \psi^{\Rightarrow x}=\psi\ll\phi\},x)\\
&=&(\phi^{\Rightarrow x},x)\\
&=&(\phi,x),
\end{array}
\end{eqnarray*}
where $e_\top=(e,\top)$.

According to the above proof, it follows that $(\Psi,D)$ is s-continuous.
\qed

\begin{theorem}\label{theorem:5}
{\rm If $(\Phi, \Phi_f,D)$ is a domain-free s-compact information algebra,
then its associated labeled information algebra is s-compact too.}
\end{theorem}
\noindent {\bf Proof.}
Let $(\Psi,D)$ be defined as in the proof of Theorem \ref{theorem:2} and
\begin{center}
$\Psi_{f,x}=\{(\psi,x):\psi\in\Phi_f,\psi=\psi^{\Rightarrow x}\}.$
\end{center}

In order to show $(\Psi,\{\Psi_{f,x}\}_{x\in D},D)$ is s-compact, by Theorem \ref{theorem:2},
it suffices to verify the axiom of compactness.
Let $\{(\psi_i,x):i\in I\}$ be a directed subset of $\Psi_{f,x}$ and $(\psi,x)\in \Psi_{f,x}$
such that $(\psi,x)\leq \bigvee\limits_{i\in I} (\psi_i,x)$. Then $\psi\leq \bigvee\limits_{i\in I} \psi_i$.
By the compactness of $(\Phi, \Phi_f,D)$, there exists a $i\in I$ such that $\psi\leq\psi_i$.
This shows that $(\psi,x)\leq(\psi_i,x)$. Hence $(\Psi,\{\Psi_{f,x}\}_{x\in D},D)$ is s-compact.
\qed

At last, we study the properties of the associated domain-free information algebra with a labeled continuous
information algebra or a labeled compact information algebra.

\begin{lemma}\cite{Lecture}\label{lemma:7}
{\rm Let $(\Phi, D)$ be a labeled information algebra, where $D$ has a top element $\top$. For $X\subseteq \Phi$,
if $\vee X\in \Phi$ exists, it holds then that
$$[\vee X]_\sigma=\vee[X]_\sigma,$$
where $[X]_\sigma=\{[\eta]_\sigma:\eta\in X\}$.}
\end{lemma}

\begin{theorem}\label{theorem:3}
{\rm If $(\Phi, D)$ is a labeled continuous information algebra,
then its associated domain-free information algebra $(\Phi/\sigma, D)$ is also continuous.}
\end{theorem}
\noindent {\bf Proof.}
1. $(\Phi/\sigma, \leq)$ is complete:
Let $\{[\eta]_\sigma:\eta\in X\}\subseteq \Phi/\sigma$. Since $\phi\equiv\phi^{ \uparrow\top}$( mod $\sigma$),
we can assume that $X\subseteq \Phi_\top$. Then $\vee X$ exists because $(\Phi_\top,\leq)$ is a complete lattice.
By Lemma \ref{lemma:7}, we have $\vee\{[\eta]_\sigma:\eta\in X\}=[\vee X]_\sigma$ .
Thus $(\Phi/\sigma, \leq)$ is a complete lattice.

2. $(\Phi/\sigma, \leq)$ is continuous:
For $[\phi]_\sigma\in \Phi/\sigma$, firstly,
we assume that $\{\psi:[\psi]_\sigma\ll[\phi]_\sigma\}\cup \{\phi\}\subseteq \Phi_\top$.
Then, by Lemma \ref{lemma:7} and the continuity in $(\Phi, D)$,
we obtain that $\vee\{[\psi]_\sigma:[\psi]_\sigma\ll[\phi]_\sigma\}=[\vee\{\psi:[\psi]_\sigma\ll[\phi]_\sigma\}]_\sigma
=[\vee\{\psi:\psi\ll_\top\phi\}]_\sigma=[\phi]_\sigma$. Then, by Theorem \ref{theorem:8},
we obtain that $(\Phi/\sigma, \leq)$ is continuous.
\qed

\begin{theorem}\label{theorem:4}
{\rm If $(\Phi, \{\Phi_{f,x}\}_{x\in D},D)$ is a labeled compact information algebra,
then its associated domain-free information algebra $(\Phi/\sigma, D)$ is also compact.

Furthermore, if $(\Phi, \{\Phi_{f,x}\}_{x\in D},D)$ is s-compact,
then its associated domain-free information algebra $(\Phi/\sigma, D)$ is s-compact too.}
\end{theorem}
\noindent {\bf Proof.}
Let $\Phi_f=\{[\phi]_\sigma: \phi\in\Phi_{f,\top}\}$. It is easy to see that $[e_\top]_\sigma\in \Phi_f$
is a neutral element of $(\Phi/\sigma, D)$. We need to verify the conditions of convergency, density, compactness and
the closeness of combination in domain-free information algebra.

1. $\Phi_f$ is closed under combination: Let $\phi,\psi\in \Phi_{f,\top}$,
then $[\phi]_\sigma\otimes [\psi]_\sigma=[\phi\otimes \psi]_\sigma\in \Phi_f$,
since $\Phi_{f,\top}$ is closed under combination.

2. Convergency:
Let $\{[\phi_i]_\sigma: i\in I\}\subseteq\Phi_f$ be directed. Then $\{\phi_i: i\in I\}\subseteq\Phi_{f, \top}$ is also directed.
By the convergency of $\Phi_{f,\top}$, $\mathop\bigvee\limits_{i\in I} \phi_i$ exists.
By Lemma \ref{lemma:7}, we have
$\mathop\bigvee\limits_{i\in I} [\phi_i]_\sigma=[\mathop\bigvee\limits_{i\in I} \phi_i]_\sigma\in \Phi/\sigma$.
Similarly, we can prove the axiom of compactness.

3. Density:
For $\phi\in\Phi$, we show $[\phi]_\sigma =\mathop\vee\{[\psi]_\sigma\in \Phi_f:[\psi]_\sigma\leq [\phi]_\sigma\}.$
In fact,
\begin{eqnarray*}
\begin{array}{lll}
 [\phi]_\sigma&=&[\phi\otimes e_\top]_\sigma\\
&=&[\mathop\vee\{\psi\in \Phi_{f,\top}:\psi\leq \phi\otimes e_\top\}]_\sigma\\
&=&\mathop\vee\{[\psi]_\sigma:\psi\in \Phi_{f,\top},\psi\leq \phi\otimes e_\top\}\\
&\leq & \mathop\vee\{[\psi]_\sigma\in \Phi_f:[\psi]_\sigma\leq [\phi]_\sigma\}\\
&\leq &[\phi]_\sigma
\end{array}
\end{eqnarray*}
Thus $[\phi]_\sigma =\mathop\vee\{[\psi]_\sigma\in \Phi_f:[\psi]_\sigma\leq [\phi]_\sigma\}$,
which implies that $(\Phi/\sigma, D)$ is compact.

If $(\Phi,\{\Phi_{f,x}\}_{x\in D},D)$ is s-compact, by the above proof, we have that $(\Phi/\sigma, D)$
is a domain-free compact information algebra.
In order to show $(\Phi/\sigma, D)$ is s-compact,
it suffices to prove the strong density in information algebra $(\Phi/\sigma, D)$.
For $\phi\in\Phi,x\in D$, by the strong density in $(\Phi, \{\Phi_{f,x}\}_{x\in D},D)$, we have
\begin{eqnarray*}
\begin{array}{lll}
 [\phi]_\sigma ^{\Rightarrow x}&=&[\phi\otimes e_\top]_\sigma^{\Rightarrow x}\\
&=&[(\phi\otimes e_\top)^{\downarrow x}]_\sigma\\
&=&[\mathop\vee\{\psi^{\downarrow x} \in \Phi_{f,x}:\psi^{\downarrow x}\otimes e_\top\in \Phi_{f,\top},
    \psi^{\downarrow x}\leq (\phi\otimes e_\top)^{\downarrow x}\}]_\sigma\\
&=&\mathop\vee\{[\psi^{\downarrow x}]_\sigma:\psi^{\downarrow x} \in \Phi_{f,x},\psi^{\downarrow x}\otimes e_\top\in
     \Phi_{f,\top}, \psi^{\downarrow x}\leq (\phi\otimes e_\top)^{\downarrow x}\}.
\end{array}
\end{eqnarray*}
Since $\psi^{\downarrow x}\otimes e_\top\in \Phi_{f,\top}$,
we have $[\psi^{\downarrow x}]_\sigma=[\psi^{\downarrow x}\otimes e_\top]_\sigma\in \Phi_f$.
Meanwhile, if $\psi^{\downarrow x}\leq (\phi\otimes e_\top)^{\downarrow x}$,
we have $[\psi^{\downarrow x}]_\sigma\leq [(\phi\otimes e_\top)^{\downarrow x}]_\sigma
=[\phi\otimes e_\top]_\sigma^{\Rightarrow x}
=[\phi]_\sigma^{\Rightarrow x}
\leq [\phi]_\sigma$.
Thus $$[\phi]_\sigma ^{\Rightarrow x}\leq
\mathop\vee\{[\psi]_\sigma\in \Phi_f:[\psi]_\sigma=[\psi]^{\Rightarrow x}_\sigma\leq [\phi]_\sigma\}
\leq [\phi]_\sigma ^{\Rightarrow x}.$$
Then $[\phi]_\sigma^{\Rightarrow x}
 =\mathop\vee\{[\psi]_\sigma\in \Phi_f:[\psi]_\sigma=[\psi]^{\Rightarrow x}_\sigma\leq [\phi]_\sigma\}.$
This proves that $(\Phi/\sigma, D)$ is s-compact.
\qed

\begin{remark}\label{remark:3}
Let us make a summary of this section.
We have proved that the associated labeled information algebra is s-continuous(resp. s-compact),
if the original domain-free information algebra is s-continuous(resp. s-compact). On the other hand,
from a labeled continuous(resp. compact/s-compact) algebra, the associated domain-free algebra we construct is also continuous
(resp. compact/s-compact).
Therefore, we can see that labeled information algebra and domain-free information algebra do correspond to each other
on s-compactness.
Unfortunately, at present, we can not obtain a conclusion that the associated domain-free information algebra with a labeled
s-continuous information algebra is s-continuous.

\end{remark}

\section{Conclusion}
In this paper, we defined the concepts of labeled continuous information algebra and domain-free continuous information
algebra. We studied the correspondence between labeled information algebra and domain-free information algebra
on continuity, s-continuity, compactness and s-compactness.
Continuous functions between two domain-free information algebras were redefined.
It has shown that function space between two domain-free s-continuous information algebras forms
a domain-free s-continuous information algebra too.

\section*{Acknowledgements}
The authors are very grateful to two anonymous referees and Professor Bernadette Bouchon-Meunier, Editor-in-Chief,
for their valuable comments and constructive suggestions which greatly improved the quality of this paper.
We are also indebted to Professor J\"{u}rg Kohlas for 
valuable and instructive suggestions. This work is partly supported
by National Science Foundation of China (Grant No.60873119) and the
Higher School Doctoral Subject Foundation of Ministry of Education
of China(Grant No.200807180005).

\end{document}